\pdfoutput=1

\documentclass[11pt]{article}

\usepackage[final]{acl}

\usepackage{times}
\usepackage{latexsym}
\usepackage{enumitem}

\usepackage[T1]{fontenc}

\usepackage[utf8]{inputenc}

\usepackage{microtype}

\usepackage{inconsolata}

\usepackage{graphicx}
\usepackage{amsmath}
\usepackage{mathtools}
\usepackage{booktabs}
\usepackage{tikz}
\usepackage{tabularx}
\usetikzlibrary{shapes,arrows,positioning}

\usepackage{tcolorbox}
\newtcolorbox{promptbox}{
    colback=gray!5!white,
    colframe=gray!50!black,
    left=2mm,
    right=2mm,
    top=1mm,
    bottom=1mm,
    fontupper=\small\ttfamily
}

\title{Incremental Summarization for Customer Support via Progressive Note-Taking and Agent Feedback}

\author{
  Yisha Wu \quad Cen (Mia) Zhao \quad Yuanpei Cao \quad Xiaoqing Xu\\
  \textbf{Yashar Mehdad} \quad \textbf{Mindy Ji} \quad \textbf{Claire Na Cheng} \\
  Airbnb Inc., USA \\
  \texttt{\{yisha.wu, mia.zhao, yuanpei.cao, xiaoqing.xu} \\
  \texttt{yashar.mehdad, mindy.ji, claire.cheng\}@airbnb.com}
}

\begin{document}
\maketitle
\begin{abstract}
We introduce an incremental summarization system for customer support agents that intelligently determines when to generate concise bullet notes during conversations, reducing agents' context-switching effort and redundant review. Our approach combines a fine-tuned Mixtral-8×7B model for continuous note generation with a DeBERTa-based classifier to filter trivial content. Agent edits refine the online notes generation and regularly inform offline model retraining, closing the agent edits feedback loop. Deployed in production, our system achieved a 3\% reduction in case handling time compared to bulk summarization (with reductions of up to 9\% in highly complex cases), alongside high agent satisfaction ratings from surveys. These results demonstrate that incremental summarization with continuous feedback effectively enhances summary quality and agent productivity at scale.
\end{abstract}

\section{Introduction}
\label{sec:intro}
Customer support agents manage complex interactions across phone, chat, and email. Agents must quickly identify core customer issues, track prior actions, and produce accurate notes. These notes are usually mandatory for handoffs, compliance, and quality audits. Industry studies show that writing these summary notes consumes roughly 10\% of case handling duration time, thus increasing context-switching effort and total working time~\citep{asapp2022contactcenter}. Two primary challenges exacerbate this issue:

\paragraph{Long multi‑source inputs} Real-world customer interactions often involve lengthy, unstructured texts and transcripts spanning thousands of tokens. As showed by DialogSum~\citep{chen2021dialogsum}, this complexity makes it difficult for agents to quickly pinpoint crucial information precisely during the interactions with customers. See Appendix~\ref{app:cs-challenges} for common challenges.

\paragraph{High accuracy demands.} Accurate summaries are essential for effective issue resolution. Any inaccuracies or omissions can lead to degraded decision-making and reduced customer satisfaction~\citep{liu2023defacto}.

\paragraph{Our Contributions} Most current summarization tools operate post-conversation without providing limited real-time support or leveraging chunk-based method to produce real-time summary based on length, which cannot address agents' context-switching effort effectively. We introduce a \emph{real-time incremental notes generation system} designed to reduce agents' context-switching effort and enhance efficiency through two key innovations: (i) a \emph{Progressive Note-Taking with Quality Control} workflow that intelligently determines optimal moments to generate concise notes using a summarization LLM and a subsequent relevance classifier, and (ii) an \emph{Agent-Edits Learning Framework}, in which agents' real-time edits immediately refine the online notes generation and periodically contribute to offline model refinement.

Deployed in production, the system reduces the average handling time by 3\%, and up to 9\% for complex scenarios, secures agent satisfaction scores over 80\%, showing that incremental summarization with live human feedback can effectively lower context-switching effort without sacrificing quality.

\section{Related Works}\label{sec:related}
\paragraph{Managing long multi-source inputs} Previous work on long customer support interactions uses hierarchical summarization and chunk-based methods. These approaches either aggregate summaries from smaller segments~\citep{li2021hierarchical}; or regenerate summaries periodically as input exceeds  thresholds. These have been extensively explored in customer service and structured meeting domains~\citep{wang2023recursive,manuvinakurike2021incremental}. While recent LLMs can process very long contexts, these generally lack precise update mechanisms needed for real-time agent support.

\paragraph{Ensuring High Accuracy} Existing work focuses on extractive and taxonomy-driven summarization, selecting key dialogue elements by speaker role, intent, or domain taxonomy to create coherent summaries~\citep{lin2021csds,swanson2010sigdial}. Human-in-the-loop optimization methods, such as reinforcement learning from human feedback and preference ranking, further improve factual precision and alignment with user expectations~\citep{chen2023hitl, stiennon2020learning}.

\paragraph{Evaluation with AI} Recent studies explore LLM-as-a-judge (LLM-judge) for automated assessment of summaries quality, showing promising correlation with expert judgments but also highlighting prompt sensitivity and bias risks~\citep{zheng2023judging,chen-etal-2024-humans}. In parallel, RL from AI feedback (RLAIF) replaces or augments human preference data with model-generated feedback, e.g., Constitutional AI, scaled RLAIF, reducing labeling cost while requiring calibration against human rubrics~\citep{bai2022constitutional,lee2023rlaif}.

Our approach extends these previous methods by introducing a production-deployed real-time summarization workflow, integrating incremental summary generation with targeted topic-based redundancy filtering, incorporates immediate edits from agents. Offline, an agent-edits learning framework leverages agent edits and an LLM-judge to curate preference pairs and automate evaluation, enabling efficient model iteration.

\section{System Architecture and Methods}
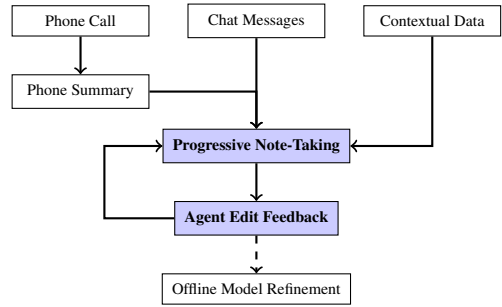
\begin{figure}[ht]
\centering
    \begin{tikzpicture}[
      node distance=0.5cm,
      box/.style={rectangle, draw, minimum width=1.8cm, minimum height=0.4cm, font=\tiny},
      arrow/.style={->, thick},
      innovation/.style={box, fill=blue!20, font=\tiny\bfseries}
    ]
    
    \node[box] (phone) {Phone Call};
    \node[box, right=0.5cm of phone] (chat) {Chat Messages};
    \node[box, right=0.5cm of chat] (context) {Contextual Data};
    
    \node[box, below=of phone] (phonesumm) {Phone Summary};
    
    \node[innovation, below=1.2cm of chat] (progressive) {Progressive Note-Taking};
    
    \node[innovation, below=of progressive] (edits) {Agent Edit Feedback};
    \node[box, below=of edits] (training) {Offline Model Refinement};
    
    \draw[arrow] (phone) -- (phonesumm);
    \draw[arrow] (phonesumm) -| (progressive);
    \draw[arrow] (chat) -- (progressive);
    \draw[arrow] (context) |- (progressive);
    \draw[arrow] (progressive) -- (edits);
    \draw[arrow, dashed] (edits) -- (training);
    
    \draw[arrow] (edits) -- +(-2,0) |- (progressive);
    
    \end{tikzpicture}
\caption{System Architecture}
\label{fig:e2e_workflow}
\end{figure}

Our unified summarization system integrates multi-channel conversations into continuously generated summary bullet notes. Agents can provide real-time edits to these notes in the UI, and their edits further refine model through periodic offline retraining (Figure~\ref{fig:e2e_workflow}).

\subsection{Progressive Note-Taking with Quality Control}\label{sec:note-taking-flow}
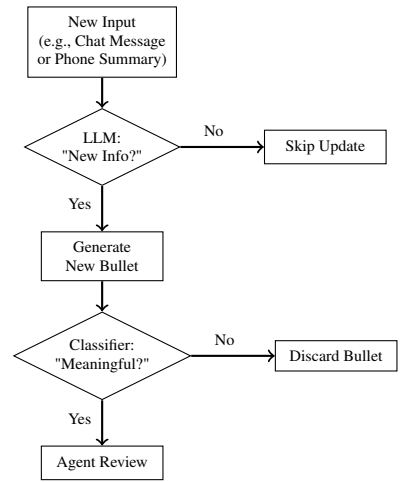
\begin{figure}[ht]
\centering
    \begin{tikzpicture}[
      node distance=0.4cm,
      box/.style={rectangle, draw, minimum width=1.6cm, minimum height=0.4cm, font=\tiny, align=center},
      decision/.style={diamond, draw, aspect=2, minimum width=1.5cm, minimum height=0.4cm, font=\tiny, align=center, inner sep=1pt},
      arrow/.style={->, thick},
      label/.style={font=\tiny}
    ]
    
    \node[box] (input) {New Input \\ (e.g., Chat Message \\ or Phone Summary)};
    \node[decision, below=of input] (llm_q) {LLM:\\"New Info?"};
    \node[box, below=of llm_q, yshift=-0.2cm] (generate) {Generate\\New Bullet}; 
    \node[decision, below=of generate] (class_q) {Classifier:\\"Meaningful?"};
    \node[box, below=of class_q, yshift=-0.2cm] (review) {Agent Review};
    
    \node[box, right=of llm_q, xshift=0.7cm] (skip) {Skip Update};
    \node[box, right=of class_q, xshift=0.7cm] (discard) {Discard Bullet};
    
    \draw[arrow] (input) -- (llm_q);
    
    \draw[arrow] (llm_q) -- node[label, left, pos=0.4] {Yes} (generate);
    \draw[arrow] (llm_q) -- node[label, above, pos=0.4] {No} (skip);
    
    \draw[arrow] (generate) -- (class_q);
    
    \draw[arrow] (class_q) -- node[label, left, pos=0.4] {Yes} (review);
    \draw[arrow] (class_q) -- node[label, above, pos=0.4] {No} (discard);
    
    \end{tikzpicture}
\caption{Progressive Note-Taking Workflow.}
\label{fig:progressive_flow}
\end{figure}

The system uses progressive note-taking, generating summary bullets only when substantial new information is detected (Figure~\ref{fig:progressive_flow}). The summarization LLM proposes candidate updates. A subsequent relevance classifier filters out trivial or redundant bullets, retaining critical updates. Agent edits provide implicit validation for summary quality.

\subsubsection{LLM-based Summarization with Continuous Bullet Generation}
We fine-tuned a Mixtral-8x7B~\citep{jiang2024mixtral} model to generate summary notes that meet the product requirements. The system operates using an iterative prompting strategy. Each invocation uses a dynamic prompt composed of task instructions, contextual data (case metadata, agent/customer profiles), full interaction history (chat, calls, emails), and previously generated and accepted bullets.

Continuous summarization uses \emph{prefix prompting} and \emph{in-context learning}. Each model response is prefixed with previous bullets, instructing the LLM to generate \textbf{only new, incremental bullets} from recent dialogue turns. If no new content is identified, the LLM returns an empty response. see Appendix~\ref{app:icn-prompts} for prompt working examples.

This iterative approach is essential, as naively re-summarizing the entire history each turn would discard agent edits to previous generated bullets. By preserving agent edit feedback, our method ensures that agent edits persist in all subsequent summary updates.

\subsubsection{Bullet Relevance Classifier for Filtering}\label{sec:icn-cls}
To improve the relevancy and conciseness of summary bullets, we fine-tuned an pretrained DeBERTa~\citep{he2020deberta} based classifier to filter out non-essential updates, such as generic acknowledgments or redundant statements. The classifier retains only high-value utterances categorized as:

\begin{itemize}[itemsep=0.5pt]
    \item \texttt{customer\_provides\_issue}: Customer describes a problem or concern.
    \item \texttt{customer\_provides\_context}: Customer shares additional details about their situation or requirements.
    \item \texttt{customer\_takes\_action}: Customer states a particular action taken or intended.
    \item \texttt{agent\_asks\_follow\_up}: Agent queries for further clarification or detail.
    \item \texttt{agent\_provides\_solution}: Agent offers a solution, recommendation, or commitment.
\end{itemize}

The bullet classifier achieved an ROC AUC of \textbf{0.96}, a macro F1 of \textbf{0.801}, and a micro F1 of \textbf{0.845}. For production, we use it as a binary classifier, retaining any of the five target classes and filtering the others, resulting in an F1 of \textbf{0.895}. Detailed fine-tuning methods and results are in Appendix~\ref{app:icn-cls-training}.

\subsection{Agent Edits Learning Framework}
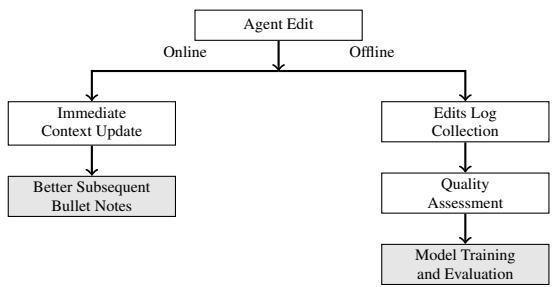
\begin{figure}[ht]
\centering
    \begin{tikzpicture}[
      node distance=0.7cm and 0.5cm, 
      box/.style={rectangle, draw, minimum width=2.2cm, minimum height=0.4cm, font=\tiny, align=center, inner sep=2pt},
      outcomebox/.style={rectangle, draw, fill=gray!20, minimum width=2.2cm, minimum height=0.4cm, font=\tiny, align=center, inner sep=2pt}, 
      arrow/.style={->, thick},
      pathlabel/.style={font=\tiny, above, yshift=1pt} 
    ]
    
        \node[box] (edit) {Agent Edit};
        
        \coordinate[below=0.4cm of edit] (branch_point);
        
        \node[box, below left=0.4cm and 1.35cm of branch_point] (context) {Immediate\\Context Update};
        \node[outcomebox, below=0.4cm of context] (notes) {Better Subsequent\\Bullet Notes}; 
        
        \node[box, below right=0.4cm and 1.35cm of branch_point] (log) {Edits Log\\Collection};
        \node[box, below=0.4cm of log] (quality) {Quality\\Assessment};
        \node[outcomebox, below=0.4cm of quality] (train) {Model Training\\and Evaluation};
        
        \draw[arrow] (edit) -- (branch_point);
        
        \draw[arrow] (branch_point) -| (context.north) node[pathlabel, pos=0.25] {Online};
        \draw[arrow] (context) -- (notes);
        
        \draw[arrow] (branch_point) -| (log.north) node[pathlabel, pos=0.25] {Offline};
        \draw[arrow] (log) -- (quality);
        \draw[arrow] (quality) -- (train);
    
    \end{tikzpicture}
\caption{Dual-path Agent Edit Feedback Workflow.}
\label{fig:agent_edit_flow}
\end{figure}

Agent edits are integral to our system's continuous improvement, operating through both online and offline mechanisms (Figure \ref{fig:agent_edit_flow}).

\subsubsection{Online Interaction Updates}
When the progressive note-taking workflow generates a new summary bullet, it appears in the agents' UI, allowing agents to correct errors, add missing details, etc. These edits are immediately saved and inform subsequent LLM prompts, ensuring the LLM continuously uses the most accurate and agent-verified information. See Appendix~\ref{app:agent-edit-prompt} how agent edits continuously update the prompt.

\subsubsection{Offline Model Refinement}\label{sec:offline-model-refinement}
To continuously improve the summarization LLM, we implement an offline pipeline that uses agent edits as valuable training signals.

All agent edits are logged as the original LLM-generated bullet ("before-edit") and the agent-modified version ("after-edit"). However, since not all edits improve quality, some may reflect individual preferences or case-specific requirements, we process these edits as follows:
\begin{itemize}
    \item \textbf{Initial Quality Assessment:} Each "after-edit" bullet is verified through a LLM-based evaluation and a sampled human review for correctness and completeness.
    \item \textbf{Conditional High-Quality Rewrite:} Bullets failing initial quality checks are rewritten by a powerful LLM, guided by verification feedback, to ensure consistent quality standards.
    \item \textbf{Pairwise Preference Data Generation:} "before-" and "after-edit" pairs are compared by LLM and human review. Only pairs with clear quality improvement are included in fine-tuning dataset, thus ensuring generalizable and meaningful training data.
\end{itemize}

Validated pairs are structured into a training corpus for further summarization LLM refinement using supervised fine-tuning (SFT) and preference alignment (DPO and ORPO~\citep{rafailov2023direct, hong2024orpo}). Detailed dataset preparation and fine-tuning methodologies are in Appendix~\ref{app:model-ft}.

\section{Evaluation Metrics}

We define explicit metrics to evaluate the summarization quality both offline and online. Offline metrics ensure precise measurement of summary quality, while online metrics measure real-world operational effectiveness.

\subsection{Offline Metrics}
\label{subsec:offline_eval_metrics}

Summary quality is assessed through conciseness, completeness, and truthfulness. Completeness and truthfulness evaluations use human reviewers and GPT-4o~\citep{hurst2024gpt} based automated LLM Judges, following the same evaluation guidelines. See details in Appendix~\ref{app:llm judge annotation prompt}.

\begin{itemize}[itemsep=0.5pt]
    \item \textbf{Conciseness:}  Measured as token reduction ratio from original input to summarized output:
    \[
    \text{Conciseness} = 1 - \frac{T_{\text{output\_summary\_tokens}}}{T_{\text{input\_conversation\_tokens}}}
    \]
    \item \textbf{Completeness:} Evaluates coverage of critical case interaction elements. Each sub-metric below is evaluated as a binary outcome (yes/no), with the overall score as their average:
    \begin{itemize}
        \item \textbf{Customer Issue:} Clearly stated customer problems.
        \item \textbf{Agent Commitment:} Explicit promises and intended follow-up actions from agents.
        \item \textbf{Agent Solution:} Key solutions and actions provided by agents.
    \end{itemize}

\item \textbf{Truthfulness:} Assesses factual accuracy compared to original interactions. Similarly evaluated as binary outcomes (yes/no) across:
\begin{itemize}
    \item \textbf{Customer Issue:} Accurate interpretation of the user's stated issue.
    \item \textbf{Agent Commitment:} Correct representation of commitments or promises made by agents.
    \item \textbf{Agent Other Action:} Precise documentation of other relevant agent activities.
\end{itemize}
\end{itemize}

\subsubsection{LLM Judge vs Human Annotators}
We compared the GPT-4o based LLM judge against traditional human annotation on 522 case samples with conversation and model generated summary notes. An expert auditor produced gold labels using the binary rubric: (i) customer issue covered, (ii) agent commitment captured, (iii) truthfulness to the raw conversation. Detailed human annotation guidelines can be found in Appendix~\ref{app:human_annotation_guideline}.

Both the LLM judge and the human annotators evaluated the model summary with the same binary rubrics. The human annotations were aggregated by majority over three passes, while the LLM judge was set at temperature 0 for stable results.

Table~\ref{tab:llm_judge_vs_human} reports each evaluator compared to the audited gold. The results indicated that the LLM judge showed superior performance, outperforming the human annotators in almost all metrics. Beyond these, the LLM judge offered a significant throughput advantage, processing evaluations approximately 15 times faster than our human annotators.

\begin{table}[ht]
\centering
\small
    \begin{tabular}{lcccc}
    \toprule
     & Accuracy & Precision & Recall & F1 \\
    \midrule
    \multicolumn{5}{c}{\textit{Customer Issue}} \\
    \midrule
    LLM Judge   & 0.974 & 0.941 & 0.593 & \textbf{0.727} \\
    Human       & 0.959 & 0.654 & 0.630 & 0.642 \\
    \midrule
    \multicolumn{5}{c}{\textit{Agent Commitment}} \\
    \midrule
    LLM Judge   & 0.965 & 0.389 & 0.583 & \textbf{0.467} \\
    Human       & 0.965 & 0.333 & 0.333 & 0.333 \\
    \midrule
    \multicolumn{5}{c}{\textit{Overall Truthfulness}} \\
    \midrule
    LLM Judge   & 0.935 & 0.410 & 0.727 & \textbf{0.525} \\
    Human       & 0.929 & 0.333 & 0.456 & 0.385 \\
    \bottomrule
    \end{tabular}
\caption{GPT-4o LLM Judge vs Human Annotators, Comparing to Gold Labels from Expert Audits}
\label{tab:llm_judge_vs_human}
\end{table}

\subsubsection{Limitations of Standard Metrics}
Standard metrics like BERTScore~\citep{zhang2019bertscore} mainly assess semantic similarity, which do not capture our core needs: factual accuracy and task-oriented utility. The complexity of customer interactions also means there is no single "gold" summary, making these metrics unreliable.

Therefore, we mainly rely on the LLM-judge-based approach for these task-oriented accuracy and effectiveness. Standard metrics like BERTScore serve only as supplementary references as shown in Appendix~\ref{app:internal-bert-score}.

\subsection{Online Metrics}\label{sec:online_eval_metrics}
To evaluate the effectiveness and impact of the system in the real world, we analyze the following online business metrics:

\begin{itemize}[itemsep=0.5pt]
    \item \textbf{Agent Satisfaction:} 
    Measured via a post-rollout survey using a five-point Likert scale (``extremely dissatisfied'' to ``extremely satisfied''), distributed to agent cohorts in English, French, and Spanish, with this new feature. Survey details are in Appendix~\ref{app:survey-details}.
    
    \item \textbf{Agent Working Time:}
    Defined as the total duration of case handling in minutes, from the initial customer interaction until case resolution.

    \item \textbf{Customer NPS (Net Promoter Score):}
    Calculated based on customer responses to standard NPS surveys sent within 14 days after case resolution. See Appendix~\ref{app:customer-nps} for details.
\end{itemize}

\section{Experiment Results}

\subsection{Background and Motivation}
Our previous system used a fine-tuned Mixtral-8x7B model from task-specific annotated data (see Appendix~\ref{app:model-ft}) to generate summaries at the end of the conversation. While it performed well compared to agent-written notes (see Appendix~\ref{app:model_vs_human}), agents reported a key limitation: they needed summaries generated \emph{incrementally during} conversations, not just at the end, to save working time.

To address this and reduce agents’ context-switching effort, we conducted a two-phase experiments to assess the impact and effectiveness of the new incremental summarization system.

\subsection{Offline Experiment}
We conducted offline experiments on 1,200 random sampled production case conversations. These experiments aimed to validate summary quality prior to production roll-out, focusing specifically on: (1) the iterative model improvements derived from agent edit feedback learning, (2) the impact of the bullet classifier to the note-taking workflow.

\subsubsection{Model Iteration Improvements}
We conducted the offline comparison to benchmark incremental improvements from our iterative fine-tuning via agent feedback learning. Evaluations were performed in three progressive model stages:

\begin{enumerate}[itemsep=0.5pt]
\item \textbf{Mixtral-Base}: The Mixtral base model without any fine-tuning.
\item \textbf{Mixtral-NF} (No Feedback): Fine-tuned variant using a task-specific dataset without incorporating agent edits, served to generate summary at the end of the case conversation.
\item \textbf{Mixtral-FB} (Feedback): Further fine-tuned from Mixtral-NF by integrating agent edit feedback, served to generate higher quality summaries for note-taking.
\end{enumerate}

We included GPT-4o as a state-of-the-art external benchmark. Models were asked to summarize the whole conversation. Summary notes were evaluated using the LLM judge approach on \textbf{completeness}, \textbf{truthfulness}, \textbf{conciseness} as mentioned in Section~\ref{subsec:offline_eval_metrics}. We also calculated the \textbf{overall} as the average of the three aspects.

\begin{table*}[ht]
    \centering
    \small
    \begin{tabularx}{\textwidth}{Xcccc}
        \toprule
         & Completeness & Truthfulness & Conciseness & Overall \\
        \midrule
        Mixtral-Base & $0.585 \pm 0.0138$ & $0.989 \pm 0.0035$ & $0.788 \pm 0.0109$ & $0.787 \pm 0.0053$\\
        Mixtral-NF & $0.824 \pm 0.0107$ & $0.995 \pm 0.0012$ & $0.819 \pm 0.0093$ & $0.879 \pm 0.0047$\\
        Mixtral-FB & $0.842 \pm 0.0011$ & $0.996 \pm 0.0019$ & $0.818 \pm 0.0095$ & $\textbf{0.885} \pm 0.0049$\\
        GPT-4o & $0.846 \pm 0.0073$ & $0.995 \pm 0.0008$ & $0.788 \pm 0.0134$ & $0.876 \pm 0.0053$\\
        \bottomrule
    \end{tabularx}
    \caption{Model Comparison Results. Scores reported as mean with 95\% CI (higher is better).}
    \label{tab:model-comp}
\end{table*}

The results in Table~\ref{tab:model-comp} showed clear incremental gains in all aspects from Mixtral-Base to the fine-tuned Mixtral-NF, and to Mixtral-FB through agent edit feedback integration. The Mixtral-FB model has the best completeness \textbf{0.842}, truthfulness \textbf{0.996}, and overall scores \textbf{0.885}. 

While GPT-4o had better completeness, its lower conciseness and operational constraints (latency, cost, data privacy) prevented its deployment (see Appendix~\ref{app:lfm-selection}). Our fine-tuned model via agent edit feedback (Mixtral-FB) data offered the optimal balance of performance and feasibility.

We also performed a paired t-test and paired bootstrap analysis with 10k replicates of the dataset, to verify the significance of performance differences between each model's overall score. The results in Table~\ref{tab:model-comp-sig} confirmed significant improvements between Mixtral-FB and Mixtral-NF or GPT-4o, while there was no significant difference between Mixtral-NF and GPT-4o.

\begin{table}[ht]
    \centering
    \small
    \begin{tabularx}{\columnwidth}{ccc}
        \toprule
        Comparison & 95\% Bootstrap CI & p-value \\
        \midrule
        Mixtral-FB vs NF        & [$+0.28\%$, $+0.96\%$] & $4.6 \times 10^{-4}$ \\
        Mixtral-FB vs GPT-4o    & [$+0.46\%$, $+1.28\%$] & $4.1 \times 10^{-5}$ \\
        Mixtral-NF vs GPT-4o    & [$-0.65\%$, $+0.13\%$] & $0.202$ \\
        \bottomrule
    \end{tabularx}
    \caption{Model Comparison Significance Results.}
    \label{tab:model-comp-sig}
\end{table}

\subsubsection{Comparative Analysis of Incremental Summarization Methods}
We conducted an ablation study on the same 1,200 case conversations to evaluate the impact of bullet classifier filtering (BCF) on progressive note-taking workflow. Each example was summarized twice using the Mixtral-FB based note-taking workflow, with and without BCF. We also compared them to the traditional chunk-based summarization methods (200- and 500-word chunks). Summaries were evaluated using the same LLM judge method as in the offline model comparison (Table~\ref{tab:icn-cls-scores}).

\begin{table*}[ht]
    \centering
    \small
    \begin{tabularx}{\textwidth}{Xcccc}
        \toprule
         & Completeness & Truthfulness & Conciseness & Overall \\
        \midrule
        Mixtral-FB Note-Taking without BCF & $0.901 \pm 0.0094$ & $0.998 \pm 0.0016$ & $0.740 \pm 0.0095$ & $0.880 \pm 0.0045$\\
        Mixtral-FB Note-Taking with BCF & $0.868 \pm 0.0135$ & $0.996 \pm 0.0022$ & $0.800 \pm 0.0078$ & $\textbf{0.888} \pm 0.0052$\\
        Mixtral-FB with 200-Word Chunks & $0.868 \pm 0.0104$ & $0.964 \pm 0.0065$ & $0.690 \pm 0.0078$ & $0.841 \pm 0.0049$\\
        Mixtral-FB with 500-Word Chunks & $0.871 \pm 0.0107$ & $0.976 \pm 0.0040$ & $0.783 \pm 0.0081$ & $0.877 \pm 0.0047$\\
        \bottomrule
    \end{tabularx}
    \caption{Incremental Summarization Ablation Comparison. Scores reported as mean with 95\% CI (higher is better).}
    \label{tab:icn-cls-scores}
\end{table*}

Although enabling the BCF slightly reduced completeness, it did not significantly affect truthfulness. The overall scores also increased from 0.880 to \textbf{0.888} due to improved conciseness. Summary length decreased by \textbf{26.5\%} on average and \textbf{24.9\%} at the median (see Appendix~\ref{app:icn-cls-training} for summary length distribution).

The chunk-based approach yielded lower truthfulness and conciseness, further supporting the effectiveness of our approach. Greater conciseness also enhanced readability and allowed agents to extract key information more efficiently, as confirmed by internal pilot feedback.

\subsubsection{Benchmark on Public Datasets}

We benchmarked Mixtral-Base, Mixtral-NF, and Mixtral-FB models on the SAMSum~\citep{gliwa2019samsum} and DialSum~\citep{chen2021dialogsum} datasets, comparing them with a few other LLM baselines. BERTScore results were reported in Table~\ref{tab:public-bert-score}.

\begin{table}[ht]
    \centering
    \small
    \begin{tabularx}{\columnwidth}{Xcc}
        \toprule
         & SAMSum & DialogSum \\
        \midrule
        Mixtral-Base    & $0.874 \pm 0.0010$ & $0.861 \pm 0.0007$ \\
        Mixtral-NF      & $0.881 \pm 0.0011$ & $0.866 \pm 0.0007$ \\
        Mixtral-FB      & $\textbf{0.888} \pm 0.0012$ & $\textbf{0.871} \pm 0.0007$ \\
        \mbox{Llama-2-ROR-FG*}  & $0.685 \pm 0.0008$ & $0.757 \pm 0.0010$ \\
        \bottomrule
    \end{tabularx}
    \caption{BERTScore F1 Comparison on Public Datasets for Reference. Results marked with * are from ~\citet{tian2024dialogue}}.
    \label{tab:public-bert-score}
\end{table}

While our refined model performed best on these public datasets, these are mainly for reference purposes. Our primary evaluation and development use a curated internal dataset, which better captures the length, complexity, and domain-specific standards critical to our application. 

\subsection{Online Experiment}

\subsubsection{Online Experiment Setup}
Due to the infeasibility of direct A/B testing caused by agent experience variance and scheduling complexities, we leveraged a Diff-in-Diff (DiD) ~\citep{abadie2005semiparametric} based quasi-experiment to evaluate our system's real-world impact.

\begin{itemize}[itemsep=0.5pt]
    \item \textbf{Treatment Group:} Pilot sites adopting the progressive note-taking workflow.
    \item \textbf{Control Group:} All other sites, which continued to use the bulk summary at the end of the conversation.
\end{itemize}

The control group agent workflow is that the agents read prior notes and conversations, chat with customer, take actions, write notes; if multiple agents are involved in the case, repeat multiple times. In the treatment group, there is no need to manually write notes, as they are automatically generated by the progressive note-taking workflow.

By comparing performance changes in both groups from three months before to one month after rollout, the DiD model isolates the feature’s true impact from external trends and site-level differences.

\subsubsection{Model Serving Configurations}\label{sec:serving_config}
The progressive note-taking workflow triggers the LLM inference at every new message or phone call. Inference is performed via NVIDIA Triton using TensorRT-LLM, with each model instance allocated 2 A100 GPUs. The p50 latency is 600 ms, and p95 is 2s.

The bullet classifier is invoked for each LLM-generated bullet. It is served using an internal model serving framework with an A10G GPU. The p50 latency is 20 ms, and p95 is 40 ms.

\subsubsection{Online Experiment Results}

\textbf{Agent Satisfaction.} Surveys from over a thousand agents showed high satisfaction with the summary feature: \textbf{95.2\%} (English), \textbf{81.8\%} (French), and \textbf{89.4\%} (Spanish) reported being "Satisfied" or "Very Satisfied".

These satisfaction scores come from the post-pilot survey targeting this feature, where agents were asked whether the summarization product benefits them.

\textbf{Agent Working Time Reduction.} Our online DiD analysis evaluated over 92\% of production cases from control and treatment sites. This revealed overall \textbf{3\%} agent working time reduction (p-val < 0.001), with up to \textbf{9\%} reduction for most complex cases (defined as cases \textbf{involving 3+ agents}, usually with solving time > 100 min).

These time savings come from: (i) model-generated notes reduced manual writing; (ii) comprehensive and easy to follow notes reduced reading raw conversations. Thus these complex cases could see a larger working time reduction. Our offline analysis showed that:

\begin{itemize}[itemsep=0.5pt]
    \item Agents' manual note-writing time ratio decreased from 10\% to 3\% after this feature roll-out.
    \item Model generated notes have higher completeness than agent's manual notes (0.871 vs 0.624 in Appendix~\ref{app:model_vs_human}).
\end{itemize}

At our production volume, these reductions correspond to \textbf{$O(10^5 - 10^6)$ agent hours saved} annually, implying \textbf{multi-million dollar cost savings} annually.

Appendix~\ref{app:icn-case-study} showed a real-world study on how our incremental summarization feature improved continuity, reduced redundancy, and minimized agent effort in complex, multi-agent, multi-channel case solving process.

\textbf{Customer NPS.} The customer NPS results remain neutral across different languages, indicating that there is no observable negative impact on customer satisfaction.

\section{Learnings}

\subsection{Learnings on Model Serving}
We evaluated optimizations in quantization and inference frameworks. 8-bit quantization reduced GPU usage (from two A100s to one) with similar latency and 1\% drop in completeness and truthfulness (tested on 2,500 cases). Furthermore, replacing vLLM with TensorRT-LLM reduced model inference latency by 20\% at p50 and p90.

The final serving stack (Section~\ref{sec:serving_config}) adopts BF16 with TensorRT-LLM, selected for the best quality in the desired latency budget, while 8-bit was retained as an efficiency fallback.

\subsection{Learnings on Multilingual Performance}
Offline evaluations revealed language-specific challenges in multilingual summarization. French summaries had lower truthfulness, mainly due to numeric inaccuracies from complex number structures, such as representing (e.g., "quatre-vingt-quatre" for 84, literally "4 × 20 + 4""). Spanish summaries had lower completeness in capturing agent solutions, due to varied verb conjugations and passive (e.g., indicative ``enviaré'', conditional ``enviaría'').

\section{Conclusions}
We introduced an incremental summarization system for customer support scenarios, leveraging progressive note-taking coupled with continuous agent feedback learning. Our approach utilizes a fine-tuned Mixtral-based LLM paired with a DeBERTa-based classifier to dynamically generate high-quality and concise summary bullets. Extensive offline evaluations validated our design, showing that iterative refinement with agent edits feedback is key for real-world alignment. Ablation studies also confirmed that the integrated bullet classifier improves summary relevance and conciseness. Real-world experiment on over 92\% of production cases showed a substantial business impact, reducing overall agent handling time by 3\%, and up to 9\% for complex cases, achieving high agent satisfaction in multi-language deployments. These confirmed the system’s effectiveness in improving operational efficiency and agent experience.

Future work will extend our system to additional languages beyond English, Spanish, and French, and incorporate multimodal contexts such as images. Additionally, we plan to further experiment with continuous retraining and updating of the classifier based on ongoing agent feedback.

\section*{Limitations}
Despite the demonstrated effectiveness of our system, several limitations remain. First, reliance on a GPT-4o-based LLM-judge for quality assessment may introduce evaluation biases or inaccuracies compared to human judgments. Second, the bullet classifier occasionally mis-classifies content, potentially causing omissions or irrelevant updates. Continuous retraining of this classifier using agent feedback will be important to improve classification accuracy and overall agent experience. Finally, summarization performance varies across different languages, and efficiently adapting our system to new languages remains a challenge for future work.

\bibliography{latex/references_v2}

\begin{thebibliography}{24}
\providecommand{\natexlab}[1]{#1}

\bibitem[{Abadie(2005)}]{abadie2005semiparametric}
Alberto Abadie. 2005.
\newblock \href {https://doi.org/10.1111/0034-6527.00321} {Semiparametric difference-in-differences estimators}.
\newblock \emph{The Review of Economic Studies}, 72(1):1--19.

\bibitem[{{ASAPP Inc.}(2024)}]{asapp2022contactcenter}
{ASAPP Inc.} 2024.
\newblock Aht meaning: What is average handle time?
\newblock \url{https://www.asapp.com/blog/aht-meaning-what-is-average-handle-time/}.
\newblock Accessed Jul 2025. Archived at \url{https://web.archive.org/web/20250704192237/https://www.asapp.com/blog/aht-meaning-what-is-average-handle-time}.

\bibitem[{Bai et~al.(2022)Bai, Kadavath, Kundu, Askell, Kernion, Jones, Chen, Goldie, Mirhoseini, McKinnon et~al.}]{bai2022constitutional}
Yuntao Bai, Saurav Kadavath, Sandipan Kundu, Amanda Askell, Jackson Kernion, Andy Jones, Anna Chen, Anna Goldie, Azalia Mirhoseini, Cameron McKinnon, and 1 others. 2022.
\newblock Constitutional ai: Harmlessness from ai feedback.
\newblock \emph{arXiv preprint arXiv:2212.08073}.

\bibitem[{Chen et~al.(2024)Chen, Chen, Liu, Jiang, and Wang}]{chen-etal-2024-humans}
Guiming~Hardy Chen, Shunian Chen, Ziche Liu, Feng Jiang, and Benyou Wang. 2024.
\newblock \href {https://doi.org/10.18653/v1/2024.emnlp-main.474} {Humans or {LLM}s as the judge? a study on judgement bias}.
\newblock In \emph{Proceedings of the 2024 Conference on Empirical Methods in Natural Language Processing}, pages 8301--8327, Miami, Florida, USA. Association for Computational Linguistics.

\bibitem[{Chen et~al.(2021)Chen, Liu, Chen, and Zhang}]{chen2021dialogsum}
Yulong Chen, Yang Liu, Liang Chen, and Yue Zhang. 2021.
\newblock \href {https://arxiv.org/abs/2105.06762} {Dialogsum: A real-life scenario dialogue summarization dataset}.
\newblock \emph{Preprint}, arXiv:2105.06762.

\bibitem[{Gliwa et~al.(2019)Gliwa, Mochol, Biesek, and Wawer}]{gliwa2019samsum}
Bogdan Gliwa, Iwona Mochol, Maciej Biesek, and Aleksander Wawer. 2019.
\newblock \href {https://doi.org/10.18653/v1/d19-5409} {Samsum corpus: A human-annotated dialogue dataset for abstractive summarization}.
\newblock In \emph{Proceedings of the 2nd Workshop on New Frontiers in Summarization}. Association for Computational Linguistics.

\bibitem[{He et~al.(2021)He, Liu, Gao, and Chen}]{he2020deberta}
Pengcheng He, Xiaodong Liu, Jianfeng Gao, and Weizhu Chen. 2021.
\newblock \href {https://arxiv.org/abs/2006.03654} {Deberta: Decoding-enhanced bert with disentangled attention}.
\newblock \emph{Preprint}, arXiv:2006.03654.

\bibitem[{Hong et~al.(2024)Hong, Lee, and Thorne}]{hong2024orpo}
Jiwoo Hong, Noah Lee, and James Thorne. 2024.
\newblock \href {https://arxiv.org/abs/2403.07691} {Orpo: Monolithic preference optimization without reference model}.
\newblock \emph{Preprint}, arXiv:2403.07691.

\bibitem[{Jiang et~al.(2024)Jiang, Sablayrolles, Roux, Mensch, Savary, Bamford, Chaplot, de~las Casas, Hanna, Bressand, Lengyel, Bour, Lample, Lavaud, Saulnier, Lachaux, Stock, Subramanian, Yang, Antoniak, Scao, Gervet, Lavril, Wang, Lacroix, and Sayed}]{jiang2024mixtral}
Albert~Q. Jiang, Alexandre Sablayrolles, Antoine Roux, Arthur Mensch, Blanche Savary, Chris Bamford, Devendra~Singh Chaplot, Diego de~las Casas, Emma~Bou Hanna, Florian Bressand, Gianna Lengyel, Guillaume Bour, Guillaume Lample, Lélio~Renard Lavaud, Lucile Saulnier, Marie-Anne Lachaux, Pierre Stock, Sandeep Subramanian, Sophia Yang, and 7 others. 2024.
\newblock \href {https://arxiv.org/abs/2401.04088} {Mixtral of experts}.
\newblock \emph{Preprint}, arXiv:2401.04088.

\bibitem[{Kennington et~al.(2023)Kennington, Fails, Wright, and Pera}]{chen2023hitl}
Casey Kennington, Jerry~Alan Fails, Katherine~Landau Wright, and Maria~Soledad Pera. 2023.
\newblock \href {https://arxiv.org/abs/2302.12043} {Conversational agents and children: Let children learn}.
\newblock \emph{Preprint}, arXiv:2302.12043.

\bibitem[{Lee et~al.(2023)Lee, Phatale, Mansoor, Mesnard, Ferret, Lu, Bishop, Hall, Carbune, Rastogi et~al.}]{lee2023rlaif}
Harrison Lee, Samrat Phatale, Hassan Mansoor, Thomas Mesnard, Johan Ferret, Kellie Lu, Colton Bishop, Ethan Hall, Victor Carbune, Abhinav Rastogi, and 1 others. 2023.
\newblock Rlaif vs. rlhf: Scaling reinforcement learning from human feedback with ai feedback.
\newblock \emph{arXiv preprint arXiv:2309.00267}.

\bibitem[{Li et~al.(2021)Li, Chen, Tung, and Chilton}]{li2021hierarchical}
Daniel Li, Thomas Chen, Albert Tung, and Lydia Chilton. 2021.
\newblock \href {https://arxiv.org/abs/2108.09597} {Hierarchical summarization for longform spoken dialog}.
\newblock \emph{Preprint}, arXiv:2108.09597.

\bibitem[{Lin et~al.(2021)Lin, Ma, Zhu, Xiang, Zhou, Zhang, and Zong}]{lin2021csds}
Haitao Lin, Liqun Ma, Junnan Zhu, Lu~Xiang, Yu~Zhou, Jiajun Zhang, and Chengqing Zong. 2021.
\newblock \href {https://arxiv.org/abs/2108.13139} {Csds: A fine-grained chinese dataset for customer service dialogue summarization}.
\newblock \emph{Preprint}, arXiv:2108.13139.

\bibitem[{Liu et~al.(2021)Liu, Sun, and Gao}]{liu2023defacto}
Yang Liu, Yifei Sun, and Vincent Gao. 2021.
\newblock \href {https://arxiv.org/abs/2106.16188} {Improving factual consistency of abstractive summarization on customer feedback}.
\newblock \emph{Preprint}, arXiv:2106.16188.

\bibitem[{Manuvinakurike et~al.(2021)Manuvinakurike, Sahay, Chen, and Nachman}]{manuvinakurike2021incremental}
Ramesh Manuvinakurike, Saurav Sahay, Wenda Chen, and Lama Nachman. 2021.
\newblock \href {https://doi.org/10.18653/v1/2021.sigdial-1.55} {Incremental temporal summarization in multi-party meetings}.
\newblock In \emph{Proceedings of the 22nd Annual Meeting of the Special Interest Group on Discourse and Dialogue}, pages 530--541, Singapore and Online. Association for Computational Linguistics.

\bibitem[{OpenAI et~al.(2024)OpenAI, :, Hurst, Lerer, Goucher, Perelman, Ramesh, Clark, Ostrow, Welihinda, Hayes, Radford, Mądry, Baker-Whitcomb, Beutel, Borzunov, Carney, Chow, Kirillov, Nichol, Paino, Renzin, Passos, Kirillov, Christakis, Conneau, Kamali, Jabri, Moyer, Tam, Crookes, Tootoochian, Tootoonchian, Kumar, Vallone, Karpathy, Braunstein, Cann, Codispoti, Galu, Kondrich, Tulloch, Mishchenko, Baek, Jiang, Pelisse, Woodford, Gosalia, Dhar, Pantuliano, Nayak, Oliver, Zoph, Ghorbani, Leimberger, Rossen, Sokolowsky, Wang, Zweig, Hoover, Samic, McGrew, Spero, Giertler, Cheng, Lightcap, Walkin, Quinn, Guarraci, Hsu, Kellogg, Eastman, Lugaresi, Wainwright, Bassin, Hudson, Chu, Nelson, Li, Shern, Conger, Barette, Voss, Ding, Lu, Zhang, Beaumont, Hallacy, Koch, Gibson, Kim, Choi, McLeavey, Hesse, Fischer, Winter, Czarnecki, Jarvis, Wei, Koumouzelis, Sherburn, Kappler, Levin, Levy, Carr, Farhi, Mely, Robinson, Sasaki, Jin, Valladares, Tsipras, Li, Nguyen, Findlay, Oiwoh, Wong, Asdar, Proehl, Yang, Antonow,
  Kramer, Peterson, Sigler, Wallace, Brevdo, Mays, Khorasani, Such, Raso, Zhang, von Lohmann, Sulit, Goh, Oden, Salmon, Starace, Brockman, Salman, Bao, Hu, Wong, Wang, Schmidt, Whitney, Jun, Kirchner, de~Oliveira~Pinto, Ren, Chang, Chung, Kivlichan, O'Connell, O'Connell, Osband, Silber, Sohl, Okuyucu, Lan, Kostrikov, Sutskever, Kanitscheider, Gulrajani, Coxon, Menick, Pachocki, Aung, Betker, Crooks, Lennon, Kiros, Leike, Park, Kwon, Phang, Teplitz, Wei, Wolfe, Chen, Harris, Varavva, Lee, Shieh, Lin, Yu, Weng, Tang, Yu, Jang, Candela, Beutler, Landers, Parish, Heidecke, Schulman, Lachman, McKay, Uesato, Ward, Kim, Huizinga, Sitkin, Kraaijeveld, Gross, Kaplan, Snyder, Achiam, Jiao, Lee, Zhuang, Harriman, Fricke, Hayashi, Singhal, Shi, Karthik, Wood, Rimbach, Hsu, Nguyen, Gu-Lemberg, Button, Liu, Howe, Muthukumar, Luther, Ahmad, Kai, Itow, Workman, Pathak, Chen, Jing, Guy, Fedus, Zhou, Mamitsuka, Weng, McCallum, Held, Ouyang, Feuvrier, Zhang, Kondraciuk, Kaiser, Hewitt, Metz, Doshi, Aflak, Simens, Boyd,
  Thompson, Dukhan, Chen, Gray, Hudnall, Zhang, Aljubeh, Litwin, Zeng, Johnson, Shetty, Gupta, Shah, Yatbaz, Yang, Zhong, Glaese, Chen, Janner, Lampe, Petrov, Wu, Wang, Fradin, Pokrass, Castro, de~Castro, Pavlov, Brundage, Wang, Khan, Murati, Bavarian, Lin, Yesildal, Soto, Gimelshein, Cone, Staudacher, Summers, LaFontaine, Chowdhury, Ryder, Stathas, Turley, Tezak, Felix, Kudige, Keskar, Deutsch, Bundick, Puckett, Nachum, Okelola, Boiko, Murk, Jaffe, Watkins, Godement, Campbell-Moore, Chao, McMillan, Belov, Su, Bak, Bakkum, Deng, Dolan, Hoeschele, Welinder, Tillet, Pronin, Tillet, Dhariwal, Yuan, Dias, Lim, Arora, Troll, Lin, Lopes, Puri, Miyara, Leike, Gaubert, Zamani, Wang, Donnelly, Honsby, Smith, Sahai, Ramchandani, Huet, Carmichael, Zellers, Chen, Chen, Nigmatullin, Cheu, Jain, Altman, Schoenholz, Toizer, Miserendino, Agarwal, Culver, Ethersmith, Gray, Grove, Metzger, Hermani, Jain, Zhao, Wu, Jomoto, Wu, Shuaiqi, Xia, Phene, Papay, Narayanan, Coffey, Lee, Hall, Balaji, Broda, Stramer, Xu, Gogineni,
  Christianson, Sanders, Patwardhan, Cunninghman, Degry, Dimson, Raoux, Shadwell, Zheng, Underwood, Markov, Sherbakov, Rubin, Stasi, Kaftan, Heywood, Peterson, Walters, Eloundou, Qi, Moeller, Monaco, Kuo, Fomenko, Chang, Zheng, Zhou, Manassra, Sheu, Zaremba, Patil, Qian, Kim, Cheng, Zhang, He, Zhang, Jin, Dai, and Malkov}]{hurst2024gpt}
OpenAI, :, Aaron Hurst, Adam Lerer, Adam~P. Goucher, Adam Perelman, Aditya Ramesh, Aidan Clark, AJ~Ostrow, Akila Welihinda, Alan Hayes, Alec Radford, Aleksander Mądry, Alex Baker-Whitcomb, Alex Beutel, Alex Borzunov, Alex Carney, Alex Chow, Alex Kirillov, and 401 others. 2024.
\newblock \href {https://arxiv.org/abs/2410.21276} {Gpt-4o system card}.
\newblock \emph{Preprint}, arXiv:2410.21276.

\bibitem[{Rafailov et~al.(2024)Rafailov, Sharma, Mitchell, Ermon, Manning, and Finn}]{rafailov2023direct}
Rafael Rafailov, Archit Sharma, Eric Mitchell, Stefano Ermon, Christopher~D. Manning, and Chelsea Finn. 2024.
\newblock \href {https://arxiv.org/abs/2305.18290} {Direct preference optimization: Your language model is secretly a reward model}.
\newblock \emph{Preprint}, arXiv:2305.18290.

\bibitem[{Stiennon et~al.(2022)Stiennon, Ouyang, Wu, Ziegler, Lowe, Voss, Radford, Amodei, and Christiano}]{stiennon2020learning}
Nisan Stiennon, Long Ouyang, Jeff Wu, Daniel~M. Ziegler, Ryan Lowe, Chelsea Voss, Alec Radford, Dario Amodei, and Paul Christiano. 2022.
\newblock \href {https://arxiv.org/abs/2009.01325} {Learning to summarize from human feedback}.
\newblock \emph{Preprint}, arXiv:2009.01325.

\bibitem[{Stolcke et~al.(2000)Stolcke, Ries, Coccaro, Shriberg, Bates, Jurafsky, Taylor, Martin, Ess-Dykema, and Meteer}]{swanson2010sigdial}
Andreas Stolcke, Klaus Ries, Noah Coccaro, Elizabeth Shriberg, Rebecca Bates, Daniel Jurafsky, Paul Taylor, Rachel Martin, Carol~Van Ess-Dykema, and Marie Meteer. 2000.
\newblock \href {https://doi.org/10.1162/089120100561737} {Dialogue act modeling for automatic tagging and recognition of conversational speech}.
\newblock \emph{Computational Linguistics}, 26(3):339–373.

\bibitem[{Tian et~al.(2024)Tian, Xia, and Song}]{tian2024dialogue}
Yuanhe Tian, Fei Xia, and Yan Song. 2024.
\newblock \href {https://doi.org/10.18653/v1/2024.acl-long.385} {Dialogue summarization with mixture of experts based on large language models}.
\newblock In \emph{Proceedings of the 62nd Annual Meeting of the Association for Computational Linguistics (Volume 1: Long Papers)}, pages 7143--7155, Bangkok, Thailand. Association for Computational Linguistics.

\bibitem[{Wang et~al.(2025)Wang, Fu, Cao, Wang, Tian, and Ding}]{wang2023recursive}
Qingyue Wang, Yanhe Fu, Yanan Cao, Shuai Wang, Zhiliang Tian, and Liang Ding. 2025.
\newblock \href {https://arxiv.org/abs/2308.15022} {Recursively summarizing enables long-term dialogue memory in large language models}.
\newblock \emph{Preprint}, arXiv:2308.15022.

\bibitem[{Wenzek et~al.(2020)Wenzek, Lachaux, Conneau, Chaudhary, Guzm{\'a}n, Joulin, and Grave}]{wenzek2020ccnet}
Guillaume Wenzek, Marie-Anne Lachaux, Alexis Conneau, Vishrav Chaudhary, Francisco Guzm{\'a}n, Armand Joulin, and {\'E}douard Grave. 2020.
\newblock \href {https://aclanthology.org/2020.lrec-1.494} {{CCNet}: Extracting high quality monolingual datasets from web crawl data}.
\newblock In \emph{Proceedings of the 12th Language Resources and Evaluation Conference (LREC)}, pages 4003--4012.

\bibitem[{Zhang et~al.(2020)Zhang, Kishore, Wu, Weinberger, and Artzi}]{zhang2019bertscore}
Tianyi Zhang, Varsha Kishore, Felix Wu, Kilian~Q. Weinberger, and Yoav Artzi. 2020.
\newblock \href {https://arxiv.org/abs/1904.09675} {Bertscore: Evaluating text generation with bert}.
\newblock \emph{Preprint}, arXiv:1904.09675.

\bibitem[{Zheng et~al.(2023)Zheng, Chiang, Sheng, Zhuang, Wu, Zhuang, Lin, Li, Li, Xing et~al.}]{zheng2023judging}
Lianmin Zheng, Wei-Lin Chiang, Ying Sheng, Siyuan Zhuang, Zhanghao Wu, Yonghao Zhuang, Zi~Lin, Zhuohan Li, Dacheng Li, Eric Xing, and 1 others. 2023.
\newblock Judging llm-as-a-judge with mt-bench and chatbot arena.
\newblock \emph{Advances in neural information processing systems}, 36:46595--46623.

\end{thebibliography}

\appendix
\section{Common Customer Support Challenges}\label{app:cs-challenges}
See Table~\ref{tab:challenges_examples} for examples.
\begin{table*}[ht]
    \centering
    \small
    \begin{tabularx}{\textwidth}{@{} l X X @{}}
        \toprule
        \textbf{Challenge} & \textbf{Raw Dialogue Snippet} & \textbf{Desired Outcome in Final Summary} \\
        \midrule
        \textbf{Integrating Disparate Sources} & 
        \textbf{Phone Call:} ``Hi, I need to cancel my booking ABC123'' \newline
        \textbf{Live Chat:} ``For my cancellation request ABC123, the reason is a last-minute family emergency.''
        & 
        \textbf Customer wanted to cancel booking ABC123 due to a family emergency. \\
        \addlinespace 
        
        \textbf{Managing Information Redundancy} & 
        \textbf{Customer:} ``My booking is ABC123.'' \newline
        \textbf{Agent:} ``Okay, I'm finding the detailed information about your booking ABC123'' \newline
        \textbf{Customer:} ``Yes, that's right, ABC123.''
        & 
        Customer confirmed booking ABC123 and agent began looking for details. \\
        \addlinespace
        
        \textbf{Maintaining Contextual Coherence} & 
        \textbf{Agent:} ``Okay, there is a cancellation fee for \$100.'' \newline
        \textbf{Customer:} ``Wait I made the booking with fully refundable price.'' \newline
        \textbf{Agent:} ``Ah sorry, you are right. You'll get the full refund.''
        & 
        Agent apologized and confirmed full refund after making mistake about cancellation fee and clarified by the customer. \\
        \bottomrule
    \end{tabularx}
    \caption{Examples of Summarization Challenges in Customer Support}
    \label{tab:challenges_examples}
\end{table*}

\section{Foundation Model Selection Comparison}\label{app:lfm-selection}
Serving cost calculation is based on 1 QPS, median input of 4k tokens and median output of 512 tokens. So the annual total input tokens are 126k million-tokens, total output tokens are 16k million-tokens. See Table~\ref{tab:model-comparison} for details.

\begin{table}[ht]
    \centering
    \small
    \begin{tabularx}{\columnwidth}{Xccc}
        \toprule
         & Model Size & Annual Cost & Latency (p95) \\
        \midrule
        GPT-4o & >100B      & \$475k/year & 4s \\
        Gemini & >100B      & \$475k/year & 17s \\
        Claude & >100B      & \$618k/year & 4s \\
        \mbox{Mixtral-8x7B} & 45B & \$100k/year & 5s \\
        \bottomrule
    \end{tabularx}
    \caption{Foundation Model Selection Comparison}
    \label{tab:model-comparison}
\end{table}

\section{Note-Taking Prompt Examples}
\subsection{Continuous Generation Examples}\label{app:icn-prompts}
Here are a few examples of how the note-taking LLM prompt works to continuously generate new bullets along with the conversation:

At the first round
\begin{promptbox}
Model Input:\\
<s>[INST] Summarize the following case conversations\\

Guest Name: Tom\\
Agent Name: Jack\\

guest(messaging): i want to refund, i cannot find the host. [/INST]\\

Model Output:\\
Guest Tom expressed his desire to request a refund but mentioned he cannot find the host
\end{promptbox}

At the second round
\begin{promptbox}
Model Input:\\
<s>[INST] Summarize the following case conversations\\

Guest Name: Tom\\
Agent Name: Jack\\

guest(messaging): i want to refund, i cannot find the host.\\
agent(messaging): i'll help you find the host. [/INST]\\

Guest Tom expressed his desire to request a refund but mentioned he cannot find the host.\\

Model output:\\
Agent Jack offered to help Tom find the host.
\end{promptbox}

At the third round
\begin{promptbox}
Model Input:\\
<s>[INST] Summarize the following case conversations\\

Guest Name: Tom\\
Agent Name: Jack\\

guest(messaging): i want to refund, i cannot find the host.\\
agent(messaging): i'll help you find the host. \\
guest(messaging): thank you [/INST]\\

Guest Tom expressed his desire to request a refund but mentioned he cannot find the host.\\
Agent Jack offered to help Tom find the host.\\

Model Output:\\
<EMPTY>\\
\end{promptbox}

\subsection{Agent Edits Prompt Examples}\label{app:agent-edit-prompt}
At the first round
\begin{promptbox}
Model Input:\\
<s>[INST] Summarize the following case conversations\\

Guest Name: Tom\\
Agent Name: Jack\\

guest(messaging): i want to refund, i cannot find the host. [/INST]\\

Model Output:\\
Guest Tom expressed his desire to request a refund but mentioned he cannot find the host
\end{promptbox}

Then agent edited the bullet to \textit{Guest Tom \textbf{wanted to} refund but he cannot find the host}. At the second round, the prompt becomes the following to reflect the agent's edits.
\begin{promptbox}
Model Input:\\
<s>[INST] Summarize the following case conversations\\

Guest Name: Tom\\
Agent Name: Jack\\

guest(messaging): i want to refund, i cannot find the host.\\
agent(messaging): i'll help you find the host. [/INST]\\

Guest Tom \textbf{wanted to} refund but he cannot find the host\\

Model output:\\
Agent Jack offered to help Tom find the host.
\end{promptbox}

\section{Bullet Classifier Training}\label{app:icn-cls-training}
The DeBERTa classifier was first pre-trained with 500G domain specific corpus mixed with CCNet dataset{~\citep{wenzek2020ccnet}}, then fine-tuned on ~70,000 bullets sourced from the production LLM summary loggings. Annotation was performed by an ensemble of LLMs like Qwen-2.5 and QwQ-32B via majority vote, guided by category definitions. The dataset was split 80:10:10 for training, validation, and testing, and the model was fine-tuned using standard cross-entropy loss.

For each category, the precision, recall, and F1 are reported in Table~\ref{tab:cls-perf}:

\begin{table}[ht]
    \centering
    \small
    \begin{tabularx}{\columnwidth}{Xccc}
        \toprule
         & Precision & Recall & F1 \\
        \midrule
        agent\_asks\_follow\_up & 0.851 & 0.850 & 0.850 \\
        agent\_provides\_solution & 0.838 & 0.879 & 0.858 \\
        customer\_provides\_context & 0.694 & 0.627 & 0.659 \\
        customer\_provides\_issue & 0.812 & 0.829 & 0.820 \\
        customer\_takes\_action & 0.714 & 0.738 & 0.726 \\
        other & 0.901 & 0.889 & 0.895 \\
        \bottomrule
    \end{tabularx}
    \caption{Bullet Classifier Per Category Performance}
    \label{tab:cls-perf}
\end{table}

\begin{figure}[ht]
    \centering
    \includegraphics[width=1\linewidth]{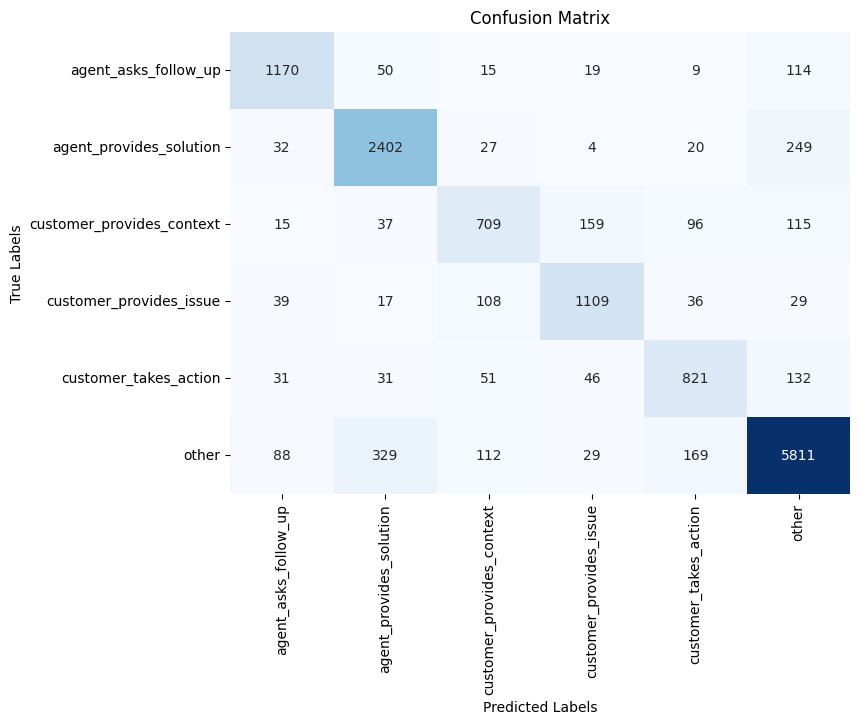}
    \caption{Confusion Matrix for the bullet Classifier}
    \label{fig:icn-confusion-matrix}
\end{figure}

\begin{figure}[ht]
    \centering
    \includegraphics[width=1\linewidth]{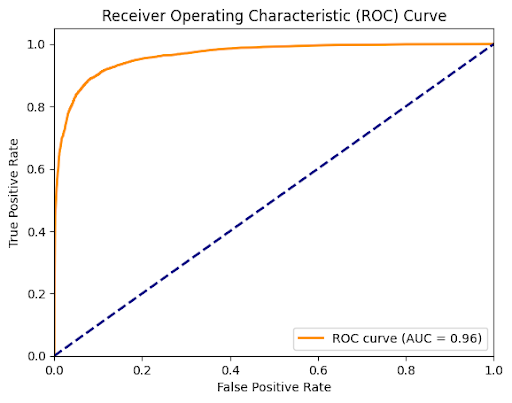}
    \caption{ROC AUC for the bullet Classifier}
    \label{fig:icn-roc-auc}
\end{figure}

See Figure~\ref{fig:icn-confusion-matrix} for the confusion matrix and Figure~\ref{fig:icn-roc-auc} for the ROC AUC curve.

\begin{figure}[ht]
    \centering
    \includegraphics[width=1\linewidth]{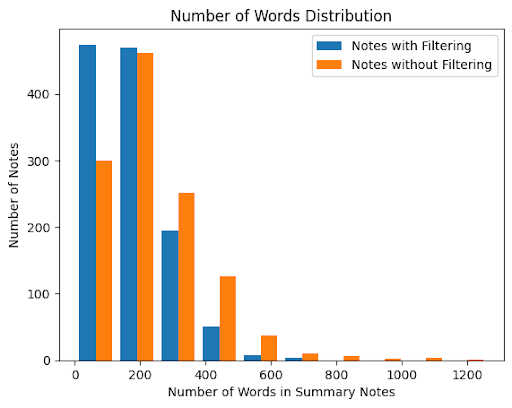}
    \caption{Distribution of Note Words Count}
    \label{fig:icn-cls-words-dist}
\end{figure}

Figure~\ref{fig:icn-cls-words-dist} showed how the bullet classifier affects the summary notes length distribution.

\section{Summarization LLM Fine-tuning Details}\label{app:model-ft}
\texttt{(prompt, chosen, rejected)} for preference alignment techniques like Direct Preference Optimization (DPO) or Odds Ratio Preference Optimization (ORPO). The \texttt{chosen} would be the high-quality "after-edit" version, and the \texttt{rejected} would be the "before-edit" version.

\texttt{(prompt, chosen)} for supervised fine-tuning (SFT), using only the high quality target.

For fine-tuning from Mixtral-Base to Mixtral-NF, we collected ~30k examples, ran 1 round of SFT and 1 round of ORPO. For fine-tuning from Mixtral-NF to Mixtral-FB, we collected ~8k examples and ran 1 round of ORPO.

\section{BERTScore Reference Metrics}\label{app:internal-bert-score}
We report BERTScore (precision, recall, F1) for our offline experiment setups on the 1,200-sample production dataset. As ground-truth summaries were unavailable, the full conversation transcript served as the reference text. We compared end-to-end summarization LLMs, and progressive note-taking workflow based on Mixtral-FB, evaluated both with and without the bullet classifier Table~\ref{tab:internal-bert-score}.

\begin{table}[ht]
    \centering
    \small
    \begin{tabularx}{\columnwidth}{Xccc}
        \toprule
         & Precision & Recall & F1 \\
        \midrule
        Mixtral-Base & 0.5083 & 0.6430 & 0.5650 \\
        Mixtral-NF & 0.5033 & 0.6475 & 0.5647 \\
        Mixtral-FB & 0.5094 & 0.6489 & 0.5691 \\
        GPT-4o & 0.5226 & 0.6418 & 0.5745 \\
        Note-Taking w/ BCF & 0.5211 & 0.6501 & 0.5763 \\
        Note-Taking w/o BCF & 0.5481 & 0.6524 & 0.5938 \\
        \bottomrule
    \end{tabularx}
    \caption{BERTScore Comparison on Internal Evaluation Dataset}
    \label{tab:internal-bert-score}
\end{table}

\section{Survey Details}\label{app:survey-details}
The examples of agent survey questions are as follows:
\begin{itemize}
    \item Model-generated summary CSAT (Customer Satisfaction Score)
    \begin{itemize}
        \item Extremely Dissatisfied
        \item Dissatisfied
        \item Neither Satisfied Nor Dissatisfied
        \item Satisfied
        \item Extremely Satisfied
    \end{itemize}
    \item How does the model-generated summary benefit you? (Multi-Select)
    \begin{itemize}
        \item Saves Time \& Efforts
        \item Provides Clear Case Context
        \item Improves Admin Notes Accuracy
        \item Captures Case Details Concisely
        \item Don't See Any Benefit
    \end{itemize}
\end{itemize}

\section{Model Summary vs Admin Notes}\label{app:model_vs_human}
Below is the human-annotated results when comparing the quality between model generated summaries and agent-written notes. Human annotators assessed each summary’s \textbf{completeness} and \textbf{truthfulness} relative to the source conversations, using the original agent notes as baselines.

\begin{table}[ht]
    \centering
    \small
    \begin{tabular}{lcccc}
        \toprule
        \textbf{Language} & \multicolumn{2}{c}{\textbf{Completeness}} & \multicolumn{2}{c}{\textbf{Truthfulness}} \\
        \cmidrule(lr){2-3}\cmidrule(lr){4-5}
        & Model & Agent & Model & Agent \\
        \midrule
        EN & 0.884 & 0.683 & 0.917 & 0.801 \\
        FR & 0.894 & 0.604 & 0.875 & 0.825 \\
        ES & 0.835 & 0.585 & 0.904 & 0.775 \\
        \bottomrule
    \end{tabular}
    \caption{Notes Quality: Model-generated Summaries vs. Agent-written Notes (higher is better).}
    \label{tab:human_vs_ai_eval_scores}
\end{table}

\section{Customer NPS (Net Promoter Score)}\label{app:customer-nps}
\textbf{NPS} is calculated as follows:
\[
\text{NPS} = \% \text{Promoters} - \% \text{Detractors}
\]

\begin{itemize}
    \item \textbf{Promoters (score 9--10):} Enthusiastic customers who are likely to recommend the company to others and demonstrate strong loyalty.
    \item \textbf{Passives (score 7--8):} Satisfied but not enthusiastic customers; unlikely to either promote or detract from the brand.
    \item \textbf{Detractors (score 0--6):} Dissatisfied customers who may harm the brand’s reputation through negative word-of-mouth.
\end{itemize}

\section{Human Annotation Guideline}\label{app:human_annotation_guideline}

\subsection{Context}

We are conducting an evaluation to assess the quality of incremental summarizations. This initiative aims to establish a baseline for the product performance while also serving as a source of truth for the development and validation of the llm judge system.

Please review both the summarized case note and the origin conversation context between the agent and the customer thoroughly and provide answers to the following questions.

\subsection{\textbf{Completeness}}

\textbf{Does the summary capture all key information about the user’s issue?}
    \begin{itemize}
        \item Yes
        \item No
        \item N/A (no such info mentioned in the conversation)
    \end{itemize}

If "no", what customer issue is missing?

Note:  
1. A summary should contain only key info necessary for a subsequent agent to effectively understand and handle the case, not all info. So, if you see that a summary doesn’t include a non-key detail that’s found in the conversation, then please still mark this as YES. Examples of non-key details: authentication action, agent's OBC-recording notification, agent actions, or negotiation process between agent and user.
2. Do not consider information that can be easily found in Atrium already, such as customer and agent name, role, or reservation details.

\textbf{Does the summary capture all the solutions provided in the conversation?}
    \begin{itemize}
        \item Yes
        \item No
        \item N/A (no such info mentioned in the conversation)
    \end{itemize}

If "no", what solution is missing?
Note: solution means that the agent resolved the user's issue (ex: 'I refunded your cleaning fee.')

\textbf{Does the summary capture all the follow-up steps provided in the conversation?}
    \begin{itemize}
        \item Yes
        \item No
        \item N/A (no such info mentioned in the conversation)
    \end{itemize}

If "no", what follow-up step is missing?

\subsection{\textbf{Accuracy}}
\textbf{Does the summary introduce any inaccurate information?}
    \begin{itemize}
        \item Yes
        \item No
    \end{itemize}

If Yes, what inaccurate information was added to the summary?

Note: Definition of accurate: This term refers to information that is correct, true, and free from errors. Accuracy is crucial as it helps establish credibility. Please only consider the accuracy of the critical information that will affect the understanding and the following handling of the case.

\textbf{Does the summary correctly interpret the commitment to the customer/promise made by the agent?}
    \begin{itemize}
        \item Yes
        \item No
        \item N/A (no such info mentioned in the summary)
    \end{itemize}

If No, what inaccurate agent commitment was added to the summary?

\textbf{Does the summary correctly interpret other agent actions that are not about the commitment/promise made?}
    \begin{itemize}
        \item Yes
        \item No
        \item N/A (no such info mentioned in the summary)
    \end{itemize}

If No, what inaccurate agent action was added to the summary?

\textbf{Does the summary correctly interpret the customer's issue?}
    \begin{itemize}
        \item Yes
        \item No
        \item N/A (no such info mentioned in the summary)
    \end{itemize}

If No, what inaccurate customer issue was added to the summary?

\textbf{Are Dates/Numbers/Currency related information correctly identified in the summary?}
    \begin{itemize}
        \item Yes
        \item No
        \item N/A (not mentioned in the summary)
    \end{itemize}

If No, what inaccurate information was added to the summary?

\subsection{\textbf{Conciseness}}
\textbf{Is the summary concise and to the point?}
    \begin{itemize}
        \item Yes
        \item No
    \end{itemize}

If No, in what ways should the summary be more concise? Give examples of lines that should be reworded for brevity.

Note: The summary should not include non-key information, such as the authentication process or negotiation process between agent and user.

\section{LLM Judge Annotation Prompt}\label{app:llm judge annotation prompt}
\subsection{Completeness}

\textbf{Evaluation of Conversation Summary Quality}

\textbf{Context:}
\begin{itemize}
    \item You are tasked with evaluating a summary generated from an Airbnb customer service conversation.
    \item The objective is to ensure the summary captures essential information.
\end{itemize}

\textbf{Instructions:}
\begin{itemize}
    \item Compare the summary with the full conversation and answer the questions below, keeping your task objective in mind.
    \item The following questions are intended to evaluate only the completeness of the information in the summary—not its accuracy. If the summary fully reflects the essential information of the conversation, answer ``Yes'', even if some details in the summary are inaccurate.
    \item Rate the summary according to the outlined criteria and provide your ratings in a JSON format as illustrated at the end.
\end{itemize}

\textbf{Completeness Questions:}
\begin{enumerate}
    \item \textbf{agent\_commitment:}
    \begin{itemize}
        \item Does the summary capture all key information about the commitment to the customer/promise made by the agent?
        \item \textbf{Answer Options:} Yes, No
        \item \textbf{Note:}
            \begin{itemize}
                \item Commitment refers to specific follow-up action that the agent promises to do after the conversation, such as calling back customers, checking information, or sending helpful links.
                \item Do not consider it a commitment if:
                \begin{itemize}
                    \item The commitment is completed during the conversation.
                    \item The commitment is just to send a recap message or documentation.
                    \item The commitment is just to wrap up and close the case or put the conversation on hold.
                    \item The commitment is offered but later not needed as the issue is resolved.
                \end{itemize}
                \item Check if all commitments promised by the agent are summarized, especially actions like sending links or special messages.
                \item This question only evaluates completeness, not truthfulness. Answer ``Yes'' if the summarized commitment is complete even if not accurate.
                \item If the summary does not mention any specific commitment or follow-up action, answer should be Yes.
            \end{itemize}
    \end{itemize}
    \item \textbf{agent\_commitment\_reason:} Provide a concise rationale for question 1 in a maximum of 50 words.
    \item \textbf{confidence\_score\_agent\_commitment:} Provide a confidence score between 0.0 and 1.0 to rate your confidence in the assessment of question 1. Use the full range (0.0 to 1.0).
    \item \textbf{confidence\_score\_agent\_commitment\_reason:} Provide a concise rationale for question 3 in a maximum of 50 words.
    \item \textbf{agent\_solution:}
    \begin{itemize}
        \item Does the summary capture all key information about the solution that agent provided to the customer?
        \item \textbf{Answer Options:} Yes, No
        \item \textbf{Note:}
            \begin{itemize}
                \item Solution refers to actions an agent takes during a conversation to resolve a customer's issue, such as confirming information, providing links, or deleting reviews.
                \item Check if all solutions provided by agent are summarized, especially actions like sending links or messages.
                \item Only evaluates completeness, not truthfulness. If the summarized solution is complete but not accurate, answer ``Yes''.
                \item If the summary does not mention any solution provided by the agent, answer should be Yes.
            \end{itemize}
    \end{itemize}
    \item \textbf{agent\_solution\_reason:} Provide a concise rationale for question 5 in a maximum of 50 words.
    \item \textbf{confidence\_score\_agent\_solution:} Provide a confidence score between 0.0 and 1.0 for question 5. Use the full range (0.0 to 1.0).
    \item \textbf{confidence\_score\_agent\_solution\_reason:} Provide a concise rationale for question 7 in a maximum of 50 words.
    \item \textbf{customer\_issue:}
    \begin{itemize}
        \item If the answer of question 1 conversation\_direction is "No", does the summary capture all key information about the customer's issue, request, and concern and explicitly state it at the beginning?
        \item \textbf{Answer Options:} Yes, No
        \item \textbf{Note:}
            \begin{itemize}
                \item Ensure the summary explicitly and directly states the customer's issue and request at the beginning.
                \item Include all key customer issues, requests, and concerns.
                \item Include all previous action taken by the customer.
                \item Include all important details (e.g., party, review concern, allergies).
                \item Only key info is necessary; minor details can be omitted.
                \item Only assesses completeness, not truthfulness.
                \item Disregard inaccuracies in confirmation codes, dates, names, or numbers.
            \end{itemize}
    \end{itemize}
    \item \textbf{customer\_issue\_reason:} Provide a concise rationale for question 9 in a maximum of 50 words.
    \item \textbf{confidence\_score\_customer\_issue:} Provide a confidence score between 0.0 and 1.0 for question 9. Use the full range (0.0 to 1.0).
    \item \textbf{confidence\_score\_customer\_issue\_reason:} Provide a concise rationale for question 11 in a maximum of 50 words.
\end{enumerate}

\subsection{\textbf{Accuracy}}
\textbf{Evaluation of Conversation Summary Quality}
\textbf{Context:}
\begin{itemize}
    \item You are tasked with evaluating a summary generated from a customer service conversation.
    \item The objective is to ensure the summary captures information accurately.
\end{itemize}

\textbf{Instructions:}
\begin{itemize}
    \item Compare the corresponding information in summary against the full conversation and answer the questions below, keeping your task objective in mind.
    \item Please note that the following question only evaluates the accuracy of the provided information, not its completeness. Therefore, do not answer "No" if there is some information missing in the summary.
    \item Rate the summary according to the outlined criteria and provide your ratings in a JSON format as illustrated at the end.
\end{itemize}
\textbf{Accuracy Questions:}
\begin{enumerate}
    \item \textbf{fake\_issue:}
    \begin{itemize}
        \item Does the summary correctly interpret the customer's issue?
        \item \textbf{Answer Options:} Yes, No
        \item \textbf{Note:} If no customer issue mentioned in summary, answer should be Yes. Disregard inaccuracies in codes, dates, names, or numbers.
    \end{itemize}
    \item \textbf{fake\_issue\_reason:} Provide a concise rationale for question 1 in a maximum of 50 words.
    \item \textbf{confidence\_score\_fake\_issue:} Provide a confidence score between 0.0 and 1.0 for question 1. Use the full range.
    \item \textbf{confidence\_score\_fake\_issue\_reason:} Provide a concise rationale for question 3 in a maximum of 50 words.
    \item \textbf{fake\_commitment:}
    \begin{itemize}
        \item Does the summary correctly interpret the commitment to the customer/promise made by the agent?
        \item \textbf{Answer Options:} Yes, No
        \item \textbf{Note:} Commitment refers to follow-up action promised. Only about truthfulness of existing commitments in summary; completeness is not assessed.
    \end{itemize}
    \item \textbf{fake\_commitment\_reason:} Provide a concise rationale for question 5 in a maximum of 50 words.
    \item \textbf{confidence\_score\_fake\_commitment:} Provide a confidence score between 0.0 and 1.0 for question 5. Use the full range.
    \item \textbf{confidence\_score\_fake\_commitment\_reason:} Provide a concise rationale for question 7 in a maximum of 50 words.
    \item \textbf{fake\_other\_action:}
    \begin{itemize}
        \item Does the summary correctly interpret other agent actions that are not about the commitment/promise made?
        \item \textbf{Answer Options:} Yes, No
        \item \textbf{Note:} If no other agent actions mentioned in transcript, answer should be Yes.
    \end{itemize}
    \item \textbf{fake\_other\_action\_reason:} Provide a concise rationale for question 9 in a maximum of 50 words.
    \item \textbf{confidence\_score\_fake\_other\_action:} Provide a confidence score between 0.0 and 1.0 for question 9. Use the full range.
    \item \textbf{confidence\_score\_fake\_other\_action\_reason:} Provide a concise rationale for question 11 in a maximum of 50 words.
    \item \textbf{role\_assignment:}
    \begin{itemize}
        \item Are the roles of all mentioned individuals (e.g., guest, host, agent) correctly identified and consistent?
        \item \textbf{Answer Options:} Yes, No
        \item \textbf{Note:} Verify all roles in the summary to ensure they align with the conversation. The focus is on accurate assignment, not correctness of names.
    \end{itemize}
    \item \textbf{role\_assignment\_reason:} Provide a concise rationale for question 13 in a maximum of 50 words.
    \item \textbf{confidence\_score\_role\_assignment:} Provide a confidence score between 0.0 and 1.0 for question 13. Use the full range.
    \item \textbf{confidence\_score\_role\_assignment\_reason:} Provide a concise rationale for question 15 in a maximum of 50 words.
    \item \textbf{fake\_digit:}
    \begin{itemize}
        \item Are Dates, Numbers, Transaction or Payment Digits, or Currency Type correctly identified in the summary?
        \item \textbf{Answer Options:} Yes, No
        \item \textbf{Note:}
            \begin{itemize}
                \item Disregard inaccuracies in confirmation codes, check-in/out dates, or codes not mentioned in the conversation.
                \item Carefully review payment amounts; cent is last two digits after decimal point.
                \item If no such information is mentioned in the summary, answer Yes.
            \end{itemize}
    \end{itemize}
    \item \textbf{fake\_digit\_reason:} Provide a concise rationale for question 17 in a maximum of 50 words.
    \item \textbf{confidence\_score\_fake\_digit:} Provide a confidence score between 0.0 and 1.0 for question 19. Use the full range.
    \item \textbf{confidence\_score\_fake\_digit\_reason:} Provide a concise rationale for question 19 in a maximum of 50 words.
\end{enumerate}

\section{Case Study for Incremental Summarization}\label{app:icn-case-study}
This case involves a guest reporting multiple concerns that resulted in escalation, handoff between multiple front-line agents, and reimbursement via the specialized team. It spanned messaging, phone and email, and involved at least 4+ agents. All names, amounts, and identifiers are redacted or anonymized.

These progressive model notes were automatically generated across the case lifecycle, making key context visible to subsequent agents without re-reading full transcripts.

\begin{promptbox}
1. Guest N reported numerous concerns ... \\
2. Guest N also mentioned aggressive wasp nest outside ... \\
3. Transferred to ... \\
4. Agent J asked N if she had any additional details to share ... \\
5. Agent J later asked N if she had any images or videos ... \\
6. Agent J suggested N could send the video via email ... \\
7. Agent I asked N to send the photo of the hotel receipt ... \\
8. Agent I informed N that the specialized team would process the reimbursement ... \\
9. Specialized team informed N that we will cover the cost ...
\end{promptbox}

\begin{itemize}
    \item After reading notes 1-3, Agent J can quickly start asking for additional details such as video evidences.
    \item After reading notes 1-6, Agent I can continue to check emails for videos and ask for receipt.
    \item After reading notes 1-8, the Specialized team can directly move to reimbursement without redundant questions.
\end{itemize}

The key takeaways are following:
\begin{itemize}
    \item Complexity: Multi-channel interaction (chat, phone, email), multiple agents, and cross-team coordination. Automated notes unify chat/phone/email events into a single evolving state, so context "travels" across agents and time.
    \item Model notes utility: Each agent continued the case without repeating prior steps or asking the guest to restate known facts.
    \item No transcript replay needed: Agents trusted and built on prior notes, reducing handling time and improving guest experience.
\end{itemize}

\end{document}